\documentclass{article}

\usepackage{neurips_2023}

\usepackage[utf8]{inputenc} % allow utf-8 input
\usepackage[T1]{fontenc}    % use 8-bit T1 fonts
\usepackage{hyperref}       % hyperlinks
\usepackage{url}            % simple URL typesetting
\usepackage{booktabs}       % professional-quality tables
\usepackage{amsfonts}       % blackboard math symbols
\usepackage{nicefrac}       % compact symbols for 1/2, etc.
\usepackage{microtype}      % microtypography
\usepackage{xcolor}         % colors

\usepackage{booktabs}
\usepackage{siunitx}
\usepackage{multirow}
\usepackage{adjustbox}
\usepackage{pifont}
\usepackage{wrapfig}
\title{Effectively Fine-tune to Improve Large Multimodal Models for Radiology Report Generation}

\author{Yuzhe Lu$^1$\thanks{Work done during an internship at Amazon} , Sungmin Hong$^2$, Yash Shah$^{2}$, Panpan Xu$^{3}$\\
$^1$Carnegie Mellon University \ \ $^2$AWS GAIIC \ \ $^3$AWS AI \\
{\texttt{yuzhelu@cs.cmu.edu, \{hsungmin, syash, xupanpan\}@amazon.com}}}

\begin{document}

\maketitle

\begin{abstract}
Writing radiology reports from medical images requires a high level of domain expertise. It is time-consuming even for trained radiologists and can be error-prone for inexperienced radiologists. It would be appealing to automate this task by leveraging generative AI, which has shown drastic progress in vision and language understanding. In particular, Large Language Models (LLM) have demonstrated impressive capabilities recently and continued to set new state-of-the-art performance on almost all natural language tasks. While many have proposed architectures to combine vision models with LLMs for multimodal tasks, few have explored practical fine-tuning strategies. In this work, we proposed a simple yet effective two-stage fine-tuning protocol to align visual features to LLM's text embedding space as soft visual prompts. Our framework with OpenLLaMA-7B achieved state-of-the-art level performance without domain-specific pretraining. Moreover, we provide detailed analyses of soft visual prompts and attention mechanisms, shedding light on future research directions.
\end{abstract}

\section{Introduction}
\label{sec:intro}

With the rapid progress in deep learning, last few years have witnessed growing interests in leveraging deep generative language models \cite{vaswani2017attention, chen2020generating, alfarghaly2021automated, nicolson2022improving} for the task of radiology report generation (RRG), whose goal is to automatically generate medical reports for an X-ray. This task is challenging as it requires 1) understanding the X-ray image fully and 2) generating clinically accurate texts. Given the impressive capabilities of LLMs on various natural language tasks, we are interested in leveraging them for the task of RRG. However, utilizing LLMs imposes unique challenges due to their huge memory requirements. Previously, the dominant strategy to combine visual and text modalities has been to add additional cross-attention layers to the language model \cite{vaswani2017attention, li2019visualbert}. This paradigm becomes infeasible for LLMs as training these additional layers demands tremendous GPU resources. As a result, many have explored the framework that uses a lightweight mapping network to project visual features to LLM's text embedding space as soft visual prompts to condition it with visual context \cite{mokady2021clipcap, chowdhery2022palm, li2023blip, van2023open, liu2023visual}. Under this framework, a common practice has been to freeze pretrained vision and language models and only train the mapping network for downstream tasks \cite{mokady2021clipcap} such as VQA \cite{van2023open}. In this work, we challenge this practice by showing that fine-tuning the vision model consistently improves performance for RRG. To avoid distorting pretrained visual features, we propose a two-stage fine-tuning strategy that firstly warms up the mapping network and demonstrate this simple strategy further improves the clinical efficacy of our model. Our contributions are summarized as below:

\begin{itemize}
    \item We demonstrate the applicability of leveraging large language models via a lightweight network mapping visual features as soft prompts for radiology report generation.
    \item We propose a simple yet effective two-stage fine-tuning strategy that improves the factuality of generated radiology reports under this framework. We show that with this fine-tuning protocol, our model provides SOTA-level performance without domain-specific pretraining. 
    \item We perform detailed analysis on the behavior of our model to reveal potential challenges when utilizing even larger language models for multimodal tasks and future directions.
\end{itemize}

\section{Related Work}
\label{sec:related}

\paragraph{Radiology Report Generation} Our research is closely related to works that use deep learning approaches to generate radiology reports from X-ray images. Many developed domain-specific methods to improve supervised fine-tuning \cite{chen2020generating, alfarghaly2021automated, nicolson2022improving}, while others explored other training paradigms such as reinforcement learning \cite{liu2019clinically, delbrouck2022improving} so as to explicitly optimize for clinical efficacy metrics. In this paper, we focus on supervised fine-tuning and note that our method is compatible with reinforcement learning techniques, which are often performed in addition to supervised fine-tuning. 

\paragraph{Large Multimodal Models} The last few years have witnessed the rise of both uni- and multi-modal foundational models \cite{bommasani2021opportunities}. These models \cite{brown2020language, radford2021learning, touvron2023llama}, pretrained on millions or even billions of data, demonstrate impressive zero-shot capabilities and can often be easily adapted for downstream tasks via fine-tuning. More recently, LLMs have shown increasingly astonishing generation capabilities as they scale up to billions of parameters. However, it is challenging for vision-language generative models to keep up with the scale as collecting paired image-text data for training is naturally much harder. Thus, many recent works have focused on combining pretrained vision models and large language models for multimodal generation tasks. In this work, we followed a popular framework of combing pretrained vision models and large language models by using a mapping network to project visual features to langauge model's text embedding space as soft visual prompts \cite{mokady2021clipcap, guo2022images, liu2023visual, li2023blip}.

\section{Methods}

In this section, we will cover the architectures, training objective, and our fine-tuning protocol to build Large Multimodal Models (LMM) from pretrained vision and large language models for RRG. 

\paragraph{Mapping Network} Our RRG model consists of three main components: a visual encoder, a mapping network, and a causal language model. The visual encoder takes in an X-ray and outputs features $\mathbf{I}$ of shape $L \times D_{img}$, where $L$ is the number of visual tokens and $D_{img}$ is the dimension of visual features. The mapping network we used is a single transformer decoder layer that takes on two roles, similar to \cite{li2023blip}. The first is to perform attention pooling \cite{lee2019set} on the visual sequence to learn a smaller number of tokens. This bottleneck structure is designed to extract features most relevant to the text modality and reduce the memory consumption of the autoregressive language model similar to \cite{li2023blip}. In our experiments, we used 10 query vectors to produce 10 compact visual tokens from the original 225 visual tokens. The second is to project visual features of $D_{img}$ to match the language model's text embedding dimension $D_{txt}$. Thus, the mapping network transforms visual features $\mathbf{I}$ of shape $L \times D_{img}$ to context $\mathbf{C}$ with $N$ tokens embeddings of dimension $D_{txt}$ that is subsequently used as soft prompts to the causal language model. 

\paragraph{Fine-tuning Objective} We fine-tune our LMM using the language modeling loss. Let $\mathbf{\theta}$ denote the parameters of our model, and $Y = \{y_1, y_2, ...y_n\}$ denote the report. The objective can be written as:
\setlength{\belowdisplayskip}{5pt}
\begin{equation}
\label{eq:lml}
\mathcal{L}(\mathbf{\theta}) = -\sum_{t=1}^{n}logp_{\mathbf{\theta}}(y_t \mid \mathbf{C}, y_{i, i < t})
\end{equation}
where $C$ is the projected visual context used to condition the language model's generation process. 

As we scale to large language models with billions of parameters, it becomes challenging to fine-tune these models end-to-end. Thus, we leverage LoRA \cite{hu2021lora}, a parameter-efficient fine-tuning method, to adapt large language models to the task of RRG. LoRA adds low-rank weight matrices to attention blocks and only trains these parameters when fine-tuning ($<1\%$ of the original parameters).

\begin{wrapfigure}{r}{0.5\textwidth}
    \begin{center}
        \includegraphics[width=0.45\textwidth]{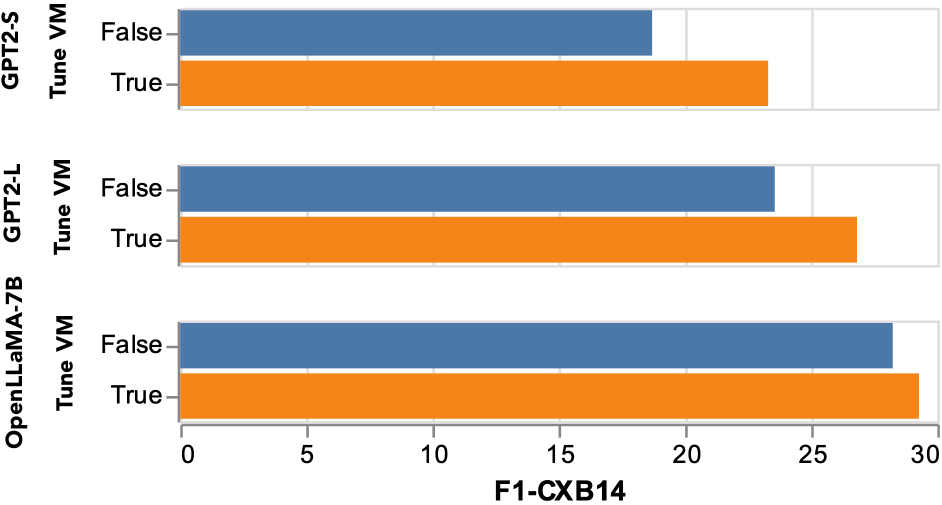}
    \end{center}
  \caption{We observed that fine-tuning the visual encoder together with the remaining model components consistently improves F1-CXB14 score.}
\label{fig:tune_vm}
\end{wrapfigure}

\paragraph{Fine-tuning Vision Encoder Improves Performance} While the common practice \cite{mokady2021clipcap, guo2022images, liu2023visual, van2023open} in recent works that combine visual and text foundational models is to freeze the visual encoder and fine-tune only the text decoder, we found fine-tuning the visual encoder together with the mapping network and text decoder consistently improves models' performance measured by clinical efficacy metrics (F1-CXB14 score), as shown in Table \ref{fig:tune_vm}. Intuitively, fine-tuning the vision encoder allows it to extract features more related to the report. However, one natural concern over fine-tuning the vision encoder is catastrophic forgetting \cite{kumar2022fine}. 

% In Figure \ref{fig:sim_matrix}, we indeed noticed that when the vision encoder is frozen, the inter-class difference is preserved better than when the vision encoder is trained. This observation leads us to propose a novel two-stage fine-tuning strategy that preserves pretrained visual features while also enjoying the benefit of end-to-end fine-tuning.

\paragraph{Two-Stage Fine-tuning Improves Performance Even Further} Following the spirit of \cite{kumar2022fine}, we proposed a similar two-stage fine-tuning strategy to avoid distorting the pretrained visual features. It is rather simple and intuitive: instead of fine-tuning the vision encoder from the very beginning, we first freeze it for one epoch to let the mapping network learn to align the two different modalities; we then unfreeze the vision encoder for the remaining epochs. We allocate only one epoch for the first stage tuning because 1) the first epoch incurs the most loss and thus most gradient updates to the vision model, and 2) we don't want extra training time. Note that we ran the same number of epochs for both one-stage and two-stage fine-tuning experiments; also, the LoRA weights are tuned in both stages since they are also randomly initialized. 

The intuition behind this two-stage protocol is that, since we have a randomly initialized mapping network, we could easily distort the visual features when we fine-tune it with the mapping network at the same time, particularly at the beginning of fine-tuning where the error of the mapping network leads to excessive updates to the vision encoder. \cite{kumar2022fine} shares the same intuition but targets the image classification setting and requires fine-tuning on downstream datasets to convergence twice, first using only linear head and then using both linear head and vision encoder. This setting requires double the training time of vanilla fine-tuning. Moreover, when we experimented with it in our problem setting, we found that the two-stage protocol in \cite{kumar2022fine} provides limited gains and can even hurt the performance (Appendix \ref{full_two_ft}). We posit that in our case, tuning the mapping network until convergence will overfit it to the language model's representations, which makes the subsequent adaptation of the vision encoder more challenging. Finally, we verify visual features after two-stage fine-tuning indeed preserve inter-class differences better than those after single-stage fine-tuning (Appendix \ref{svp}).  

\begin{table*}[]
    \centering
    \begin{adjustbox}{width=\textwidth}
    \begin{tabular}{c|c|c|c|c|c|c|c|c}
    \toprule
        \multirow{2}{*}{\textbf{Method}} & \multirow{2}{*}{\textbf{LM}} & \multirow{2}{*}{\textbf{LM Total}} & \multirow{2}{*}{\textbf{LM Tunable}} & \multirow{2}{*}{\textbf{2-Stage}} &  \multicolumn{4}{c}{\textbf{MIMIC-CXR}}\\
         & & & & & BLEU4 & ROUGE-L & F1-CXB-14 & F1-CXB-5 \\
         \cmidrule{1-9}
         \textbf{BioVIL-T} & CXRBERT & 138M & 138M & N/A & \textbf{6.9} & 23.1 & 31.4 & 42.0\\
         \cmidrule{1-9}
         \multirow{6}{*}{\textbf{Ours}} & \multirow{2}{*}{GPT2-S} & \multirow{2}{*}{117M} & \multirow{2}{*}{0.29M} & \ding{55} &  \underline{6.4} & \underline{23.2} & 23.3 & \underline{31.2} \\
         \quad & \quad & \quad & \quad & \ding{51} & 5.8 & 22.6 & \underline{24.6} & 30.6 \\
         \cmidrule{2-9}
         \quad & \multirow{2}{*}{GPT2-L} & \multirow{2}{*}{774M} & \multirow{2}{*}{1.47M} & \ding{55} & 6.3 & 23.2 & 26.8 & 37.7 \\
         \quad & \quad & \quad & \quad & \ding{51} & \underline{6.5} & \textbf{\underline{23.8}} & \underline{28.4} & \underline{39.6} \\
         \cmidrule{2-9}
         \quad & \multirow{2}{*}{OpenLLaMA} & \multirow{2}{*}{7B} & \multirow{2}{*}{4.19M} & \ding{55} & 6.8 & 23.3 & 29.3 & 42.0 \\
         \quad & \quad & \quad & \quad & \ding{51} & \textbf{\underline{6.9}} & \underline{23.5} & \textbf{\underline{32.0}} & \textbf{\underline{42.2}} \\
         \cmidrule{1-9}
    \bottomrule
    \end{tabular}
    \end{adjustbox}
    \caption{We observe consistent performance gains when utilizing larger language models and the two-stage fine-tuning protocol. Our best-performing model with OpenLLaMA-7B achieves SOTA-level performance without extensive pretraining on the MIMIC-CXR dataset as in BioVIL-T \cite{boecking2022making}.}
    \label{tab:main}
    \vspace{-5pt}
\end{table*}

\section{Experiments and Results}

In this section, we cover our experimental setup and provide analysis our results. 

\paragraph{Datasets} We used the MIMIC-CXR dataset \cite{johnson2019mimic} following preprocessing from \cite{delbrouck-etal-2022-vilmedic} . Since we aimed to generate the \textit{findings} section of the report, we disregarded ones with a missing \textit{findings} section. As most images have frontal views, we followed \cite{bannur2023learning} and only used AP/PA views. In practice, radiologists frequently refer to previous images when drafting reports for the current X-ray. Training models using these reports without providing previous images encourages hallucinating a non-existent prior. To focus on describing the details in the image, we followed the spirit of \cite{ramesh2022improving} and curated a subset that doesn't use additional context other than the current image. Specifically, we used information from the \textit{comparison} section from the radiology report and filtered out reports whose comparison section explicitly states previous sessions were referred to. After this step, we have curated a cross-sectional dataset consisting of 85,802 training, 682 validation, and 1259 test data, which is roughly half of the original dataset, and each comprising of a single image and report pair.

\paragraph{Vision and Language Models} For the visual encoder, we utilized the same vision encoder $E_{img}$ as \cite{boecking2022making}, a pretrained ResNet50. We chose it since this model was pretrained on high-resolution ($480 \times 480$) chest X-rays and presumably learned more discriminative representations. For the language model,  we utilized three models of different scales, GPT2-S (117M), GPT2-L (774M) \cite{radford2019language}, and OpenLLaMA-7B (7B) \cite{openlm2023openllama}. We include the fine-tuning details in Appendix \ref{imp_detail}. 

% One important distinction between our setting and BioVIL-T is that we utilize visual features $\mathbf{I}$ as soft prompts while BioVIL-T incorporates visual features through cross-attention layer in every transformer block in its language model. Training these cross-attention layers would require significantly more memories that make it unfeasible to fine-tune billion-parameter LLMs on consumer GPUs. 

% For the language model, BioVIL-T used a causal CXR-BERT model \cite{boecking2022making} (138M). In our experiments, we utilized three models of different scales, GPT2-S (117M), GPT2-L (774M) \cite{radford2019language}, and OpenLLaMA-7B (7B) \cite{openlm2023openllama}. One important distinction between our setting and BioVIL-T is that we utilize visual features $\mathbf{I}$ as soft prompts while BioVIL-T incorporates visual features through cross-attention layer in every transformer block in its language model. Training these cross-attention layers would require significantly more memories that make it unfeasible to fine-tune billion-parameter LLMs on consumer GPUs. 

\paragraph{Evaluation Metrics} To evaluate the quality of generated radiology reports, we use both Natural Language Generation (NLG) metrics and Clinical Efficacy (CE) metrics. For NLG Metrics, we report BLEU4 \cite{papineni2002bleu} and ROUGE-L \cite{lin-2004-rouge}. To evaluate the factual completeness and correctness of the generated report, we used the F1CheXBert score, which uses CheXBert \cite{smit2020chexbert}, a deep learning 
 based chest radiology labeler that outputs 14 relevant medical observations from a given report. F1CheXBert is essentially the F1 score between the output of CheXBert on generated report $\hat{y}$ and ground truth report $y$. To make the comparison to prior works easier, we also report F1CheXBert for the 5 most prevalent observations. We denote these metrics as F1-CXB-14 and F1-CXB-5 respectively. 

\paragraph{Main Results} In Table \ref{tab:main}, we show our main experimental results. We observed consistent performance gains on both NLG and CE metrics as we scaled our language model from GPT2-S to OpenLLaMA-7B. Moreover, we found that our two-stage fine-tuning strategy effectively improved F1-CXB14 scores for all models, with an average improvement of 1.9 points. To contextualize our models' performance, we reproduced a SOTA method, BioVIL-T \cite{boecking2022making}. Despite BioVIL-T having access to 2x more data from MIMIC-CXR for task-specific pretraining, we found our best-performing model with OpenLLaMA-7B and two-stage fine-tuning manages to outperform it. We believe that our framework of leveraging off-the-shelf LMMs provides promising results for the task of radiology report generation. We provide a qualitative evaluation in Appendix \ref{qual_results}.

\begin{figure*}[ht]
  {\includegraphics[width=\linewidth]{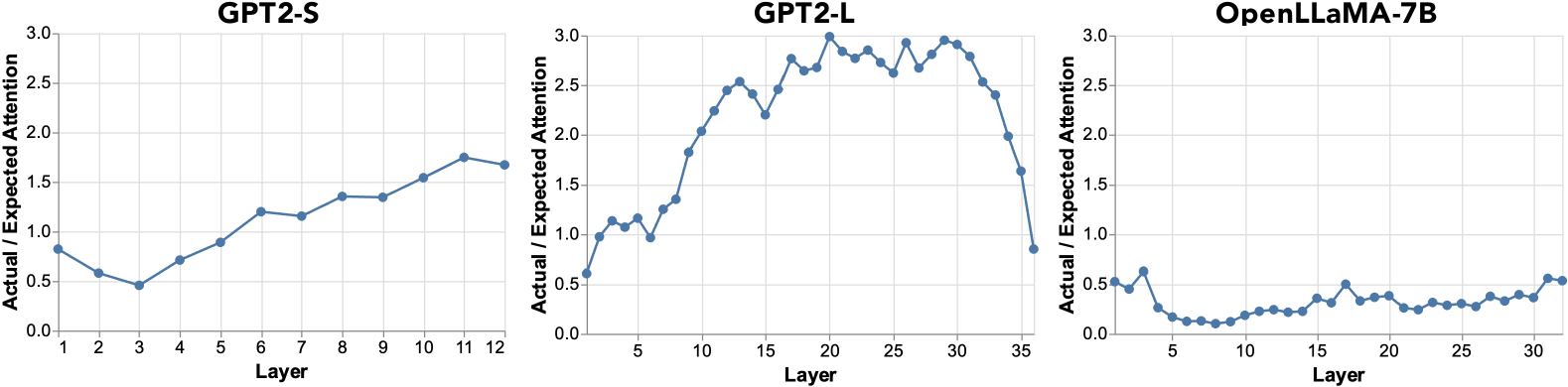}}
  \caption{We visualize the average attention weights allocated to the visual prompts across all attention layers. One critical observation is that OpenLLaMA-7B gives much less attention to soft visual prompts than smaller GPT2 models. This suggests that OpenLLaMA-7B, and potentially larger language models in general, might be less grounded to visual information.}
  \vspace{-5pt}
\label{fig:att}
\end{figure*}

\paragraph{Attention Allocation} While we are able to outperform BioVIL-T with OpenLLaMA-7B, we admit the performance gain when scaling from GPT2-L to OpenLLaMA-7B is smaller than we expected. Meanwhile, we observed OpenLLaMA-7B had a much lower validation loss than BioVIL-T, despite their similar test performances. We hypothesize OpenLLaMA-7B might overly rely on its language modeling ability and pay less attention to the soft visual prompts. To verify this hypothesis, we visualized the attention to the visual soft prompts for three models in Figure \ref{fig:att}. Concretely, each newly generated token has an attention matrix $O$ of shape $H \times L$ at each layer, where $H$ is the number of heads and $L$ is the number of generated tokens. As the length of generated tokens grows, absolute attention weights allocated to the visual tokens naturally get diluted. Thus, we adjusted $O\prime$ using a multiplier $L / K$, where K is the length of visual prompts. Finally, we compute the amount of attention allocated to visual prompts in $O\prime$ by averaging across attention heads and test samples. Thus, an average attention of 1 means the visual prompt gets attention proportional to its length. 
\setlength{\belowdisplayskip}{5pt}
\begin{equation}
\alpha_i = {\sum_{j=1}^{K} attention[i][j]} \:\big/\: {\frac{K}{K + i}}, \quad i \in {1, ...n}
\end{equation}
Interestingly, we found that OpenLLaMA-7B pays much less attention to the visual prompt on average compared to GPT2-S and GPT2-L, validating our previous hypothesis. Intuitively, with its teacher-forcing scheme, the language modeling loss \ref{eq:lml} can be too easy for LLMs with billions of parameters. Essentially, since LLM is too good at predicting the next word, it becomes hard to align visual tokens to LLM's text embedding space with the loss alone. In fact, this is also implicitly confirmed in Figure \ref{fig:sim_matrix}, where we can see that the average pair-wise similarity between test samples is higher overall than GPT2-L (darker color means lower similarity). This serves as evidence that the language modeling loss did not effectively force the mapping network of OpenLLaMA to separate the visual features of samples from different classes.

\paragraph{Conclusion and Future Work} In this paper, we utilized a lightweight framework to build large multimodal models from off-the-shelf vision and large language models for the task of RRG. To maximize the performance gains, we proposed two-stage fine-tuning to align visual features to LLM's embedding space as soft prompts and demonstrated this strategy consistently improves clinical accuracy for language models across different scales. By visualizing the attention weights, we found the soft visual prompt doesn't receive consistent attention, especially when using larger language models. This could make the generated report less grounded. In the future, we plan to investigate more sophisticated ways of incorporating visual features into LLMs. We think visual-conditioned prefixing tuning \cite{li2021prefix} can be a promising solution. 

% Finally, we found that the average confidence of current models are not reflective of their factuality. Future works could focus on finding better uncertainty metrics or designing calibration methods for generation tasks. 

% \begin{ack}
% Use unnumbered first level headings for the acknowledgments. All acknowledgments go at the end of the paper before the list of references.

% Do {\bf not} include this section in the anonymized submission, only in the final paper. You can use the \texttt{ack} environment provided in the style file to automatically hide this section in the anonymized submission.
% \end{ack}

% \bibliographystyle{neurips_2023}
\bibliographystyle{plain}
\bibliography{reference}

\newpage
\appendix

\section{Implementation Details} 
\label{imp_detail}
We used AdamW optimizer \cite{loshchilov2017decoupled} for supervised fine-tuning. We train our vision encoder and mapping network end-to-end while applying LoRA to our language model. Specifically, we added low-rank matrices of dimension 8 to the query $Q$ and value $V$ in each transformer block. We also used gradient-checkpointing \cite{chen2016training} to further reduce memory consumption. With these techniques, we were able to use a batch size of 16 for all of our experiments. For learning rate, we experimented with $[2\mathrm{e}{-5}, 5\mathrm{e}{-5}, 1\mathrm{e}{-4}]$ and selected the best learning rate by best F1-CXB-14 score on validation set. We found $1\mathrm{e}{-4}$ works best for GPT2-S and $5\mathrm{e}{-5}$ works best for GPT2-L and OpenLLaMA-7B when freezing the vision encoder and used the same learning rates for other experiments with the same language model. We trained all models for 50 epochs. We used the checkpoint with the best validation F1-CXB-14 score for testing. In terms of hardware, we were able to fine-tune the OpenLlama-7B model on a single 24G GPU instance while also fully fine-tuning our vision encoder. 

\section{Soft Visual Prompts}
\label{svp}
To verify our intuition that two-stage fine-tuning helps to preserve pretrained visual features, we propose to analyze the soft visual prompts of input images. Our idea is that if pretrained features are preserved better, difference between positive sample pairs’ similarity scores and negative sample pairs’ similarity scores should be larger on average, where we define positive pairs as samples with the same labels, and negative pairs as samples with different labels. 

\begin{figure}[ht]
   \centering
  {\includegraphics[width=0.7\linewidth]{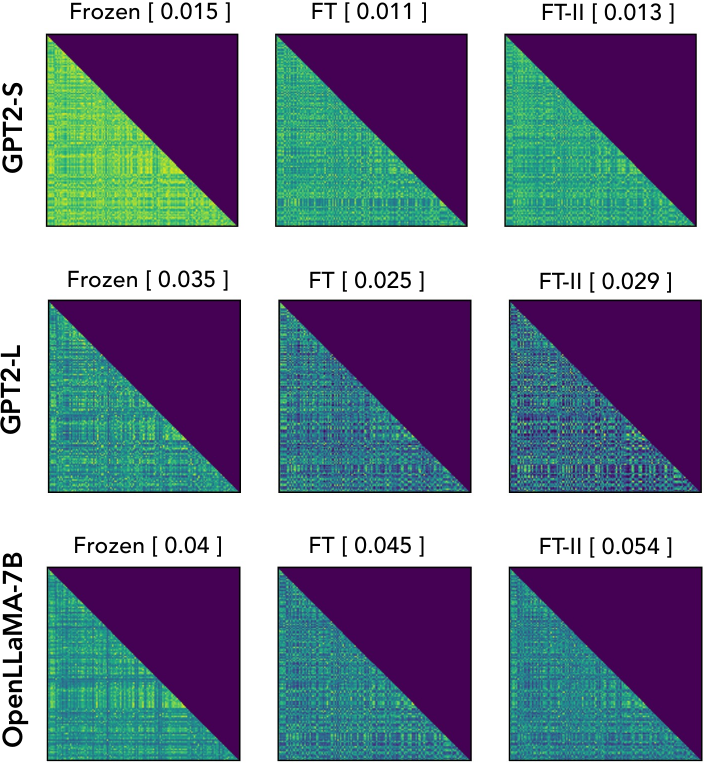}}
  {\caption{Pairwise cosine similarity between test samples' visual soft prompts. Test samples are sorted based on their ground truth labels. The number in square bracket denotes the relative difference between mean cosine similarity between positive pairs and that of negative pairs as detailed in Eq. \ref{eq:svp}. Brighter color denotes higher cosine similarity. Here, Frozen means vision model is not tuned, FT denotes vanilla fine-tuning, and FT-II denotes two-stage fine-tuning.}}
\label{fig:sim_matrix}
\end{figure}

Each visual prompt consists of $K$ visual tokens of dimension $D_{text}$. To compute similarity between visual prompts, we flatten it into a long vector of dimension $K \times D_{text}$. To quantify the difference between positive pairs and negative pairs on average, we propose to use the following metric:  

\begin{equation}
\label{eq:svp}
\Delta = \frac{\frac{1}{|P|}\sum\limits_{i, j \in P} cos(F_i, F_j)}{\frac{1}{|N|}\sum\limits_{p, q \in N} cos(F_p, F_q)} - 1
\end{equation}

where $P$ denotes the set of positive sample pairs while $N$ denotes the set of negative sample pairs. 

We present our analysis in Fig \ref{fig:sim_matrix}. Each subplot is a cosine similarity matrix of all test samples' soft visual prompts. We sort test samples by their labels so that clusters of positive pairs are better visualized. We make two observations from these plots. Firstly, tuning vision model makes the visual prompts more distinguishable in general, as evidenced the smaller cosine similarity scores on average. Secondly, two-stage fine-tuning preserves the similarity between positive pairs better, as evidenced by the consistently higher $\Delta$ value of two-stage fine-tuning than vanilla fine-tuning in Fig \ref{fig:sim_matrix}.

\section{Additional Results on Two-Stage Fine-tuning}
\label{full_two_ft}

In addition to our proposed two-stage fine-tuning setting, we also experimented with the setting where we fine-tune the model for the full amount of epochs for each stage, similar to \cite{kumar2022fine} and showed the results in Table \ref{tab:two_stage}. We observed mixed results between the two in terms of clinical efficacy scores. We think our proposed setting is more desirable since it doesn't require additional training epochs.

\section{Qualitative Results}
\label{qual_results}
\begin{figure*}[ht]
  {\includegraphics[width=\linewidth]{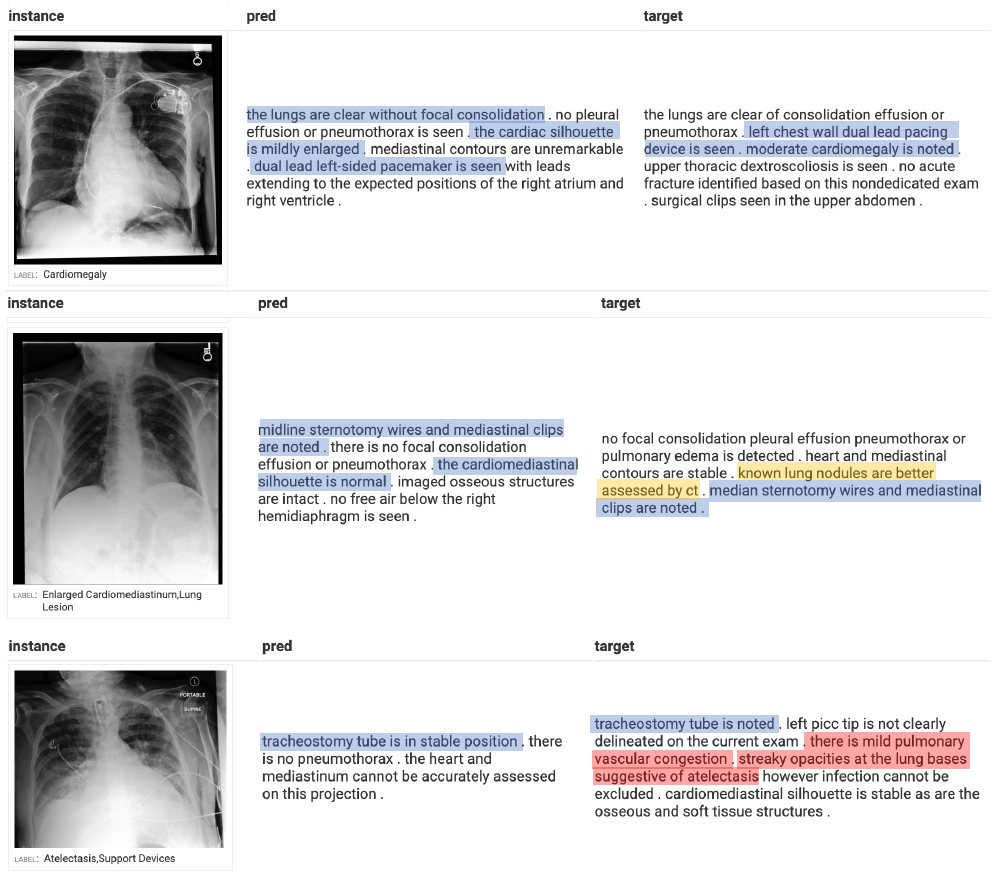}}
  {\caption{The first row showcases a case where the model accurately identifies cardiomegaly (enlarged heart). The second row demonstrates an interesting case where a medical observation is mentioned according to outside knowledge but not well observed in the current X-ray, which renders the miss acceptable. The third row shows a failure case where our model fails to capture critical medical observations.  }}
\label{fig:examples}
\end{figure*}

We provide several examples of generated reports from our OpenLLaMA model in Fig. \ref{fig:examples}, using Zeno \cite{cabrera2023zeno}.

\begin{table*}[]
    \centering
    \begin{adjustbox}{width=\textwidth}
    \begin{tabular}{c|c|c|c|c|c|c|c|c}
    \toprule
        \multirow{2}{*}{\textbf{Method}} & \multirow{2}{*}{\textbf{LM}} & \multirow{2}{*}{\textbf{LM Total}} & \multirow{2}{*}{\textbf{LM Tunable}} & \multirow{2}{*}{\textbf{Stage-1 Epochs}} &  \multicolumn{4}{c}{\textbf{MIMIC-CXR}}\\
         & & & & & BLEU4 & ROUGE-L & F1-CXB-14 & F1-CXB-5 \\
         \cmidrule{1-9}
         \multirow{6}{*}{\textbf{Ours}} & \multirow{2}{*}{GPT2-S} & \multirow{2}{*}{117M} & \multirow{2}{*}{0.29M} & 1 &  5.8 & 22.6 & \underline{24.6} & 30.6 \\
         \quad & \quad & \quad & \quad & 50 & \underline{5.9} & \underline{23.1} & 24.1 & \underline{34.5} \\
         \cmidrule{2-9}
         \quad & \multirow{2}{*}{GPT2-L} & \multirow{2}{*}{774M} & \multirow{2}{*}{1.47M} & 1 & 6.5 & 23.8 & 28.4 & \underline{39.6} \\
         \quad & \quad & \quad & \quad & 50 & \underline{7.1} & \underline{24.0} & \underline{28.8} & 38.1 \\
         \cmidrule{2-9}
         \quad & \multirow{2}{*}{OpenLLaMA} & \multirow{2}{*}{7B} & \multirow{2}{*}{4.19M} & 1 & \underline{6.9} & \underline{23.5} & \underline{32.0} & 42.2 \\
         \quad & \quad & \quad & \quad & 50 & 6.41 & 23.0 & 30.5 & \underline{42.3} \\
         \cmidrule{1-9}
    \bottomrule
    \end{tabular}
    \end{adjustbox}
    \caption{We show the results for two slightly different experimental settings of two-stage fine-tuning. }
    \label{tab:two_stage}
\end{table*}

\section{Confidence} 
\begin{figure*}[ht]
  {\includegraphics[width=\linewidth]{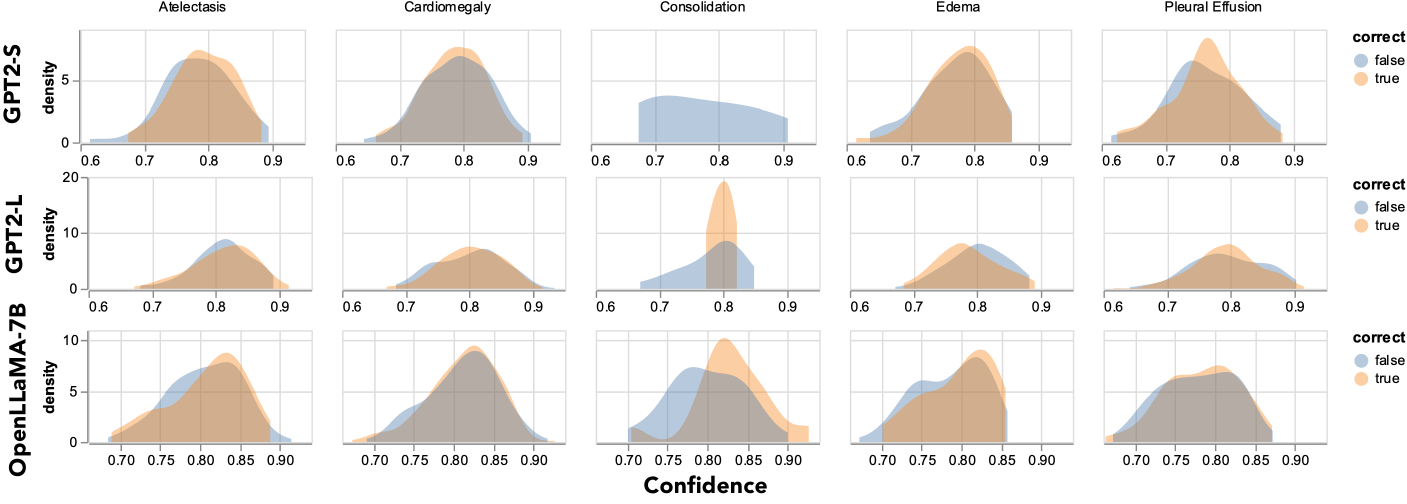}}
  {\caption{Distribution of confidence scores for true and positive cases.}}
\label{fig:conf_dist}
\end{figure*}

Additionally, we analyzed the confidence scores of generated reports, which is less explored in previous works. Since a common issue of LLMs is hallucination, we investigate if the model is aware of hallucination (defined as false positive cases) by looking at its confidence scores for true positive and false positive cases respectively. We provide a starting point for analyzing model confidence distribution by considering the average confidence $\textit{AC}$ of a generated report $Y = \{y_1, y_2, ...y_n\}$, defined as the following:

\begin{equation}
{AC}(\mathbf{Y} \mid \mathbf{\theta}) = \frac{1}{n}\sum_{t=1}^{n}p_{\mathbf{\theta}}(y_t \mid \mathbf{C}, y_{i, i < t})
\end{equation}

where $\mathbf{\theta}$ is the model parameter and $\mathbf{C}$ is the visual context. In Figure \ref{fig:conf_dist}, we provide a visualization of average confidence distribution grouped by classes for all three of our models. We found that the AC distribution of true positives and false positives tend to have overlapping supports. This unfortunately makes it extremely challenging to signal potential mistakes to the end user.

% \begin{figure}[]
% \centering
%   {\includegraphics[width=0.7\textwidth]{Figures/conf_diff.pdf}}
%   {\caption{The mean confidence for true positive and false positive cases on the test set for our models.}}
%   \label{fig:conf}
% \end{figure}

\end{document}